\documentclass[lettersize,journal]{IEEEtran}
\usepackage{amsmath,amsfonts}
\usepackage{algorithmic}
\usepackage{algorithm}
\usepackage{array}
\usepackage[caption=false,font=normalsize,labelfont=sf,textfont=sf]{subfig}
\usepackage{textcomp}
\usepackage{stfloats}
\usepackage{url}
\usepackage{verbatim}
\usepackage{graphicx}
\usepackage{cite}
\usepackage{graphicx}
\usepackage{amssymb}
\usepackage{booktabs}
\usepackage{color}
\usepackage{lettrine}
\usepackage{mathrsfs}
\usepackage{amsmath}
\usepackage{amsfonts}
\usepackage{xr}
\usepackage{bm}
\usepackage{multirow}
\usepackage[switch]{lineno} 
\usepackage{amsfonts,amssymb}
\usepackage{booktabs}
\usepackage{bbding}
\usepackage{algorithm}
\usepackage{algorithmic}
\usepackage{times}  
\usepackage{helvet}  
\usepackage{courier}  
\usepackage{gensymb}
\usepackage{tikz}
\usepackage{pgfplots}
\usepackage{stfloats}
\usepackage{url}
\usepackage{verbatim}
\usepackage{graphicx}
\usepackage{cite}

\usepackage{color}
\definecolor{turquoise}{cmyk}{0.65,0,0.1,0.3}
\definecolor{purple}{rgb}{0.65,0,0.65}
\definecolor{dark_green}{rgb}{0, 0.5, 0}
\definecolor{orange}{rgb}{0.8, 0.6, 0.2}
\definecolor{red}{rgb}{0.8, 0.2, 0.2}
\definecolor{darkred}{rgb}{0.6, 0.1, 0.05}
\definecolor{blueish}{rgb}{0.0, 0.3, .6}
\definecolor{light_gray}{rgb}{0.7, 0.7, .7}
\definecolor{pink}{rgb}{1, 0, 1}
\definecolor{greyblue}{rgb}{0.25, 0.25, 1}

\newcommand{\yz}[1]{{\color{black}{#1}}}

\begin{document}

\title{Surround-view Fisheye BEV-Perception for Valet Parking: Dataset, Baseline and Distortion-insensitive Multi-task Framework}

\author{Zizhang Wu$^{1\&}$ \and Yuanzhu Gan$^{1\&}$ \and Xianzhi Li$^{2*}$ \and Yunzhe Wu$^{1}$ \and Xiaoquan Wang$^{1}$ \and Tianhao Xu$^{3}$  \and Fan Wang$^{1}$
\thanks{1 Zongmu Technology}
\thanks{2 Huazhong University of Science and Technology}
\thanks{3 Technical University of Braunschweig}
\thanks{\& These authors contributed equally to this work and should be considered co-first authors.}
\thanks{$\ast$ \ Corresponding author: xzli@hust.edu.cn(X. Li)}}



\maketitle

\begin{abstract}
Surround-view fisheye perception under valet parking scenes is fundamental and crucial in autonomous driving.
Environmental conditions in parking lots perform differently from the common public datasets, such as imperfect light and opacity, which substantially impacts on perception performance.
Most existing networks based on public datasets may generalize suboptimal results on these valet parking scenes, also affected by the fisheye distortion.
In this article, we introduce a new large-scale fisheye dataset called \textbf{F}isheye \textbf{P}arking \textbf{D}ataset (\textbf{FPD}) to promote the research in dealing with diverse real-world surround-view parking cases. 
Notably, our compiled \textbf{FPD} exhibits excellent characteristics for different surround-view perception tasks.
In addition, we also propose our real-time distortion-insensitive
multi-task framework \textbf{F}isheye \textbf{P}erception \textbf{Net}work (\textbf{FPNet}), which improves the surround-view fisheye BEV perception by enhancing the fisheye distortion operation and multi-task lightweight designs. 
Extensive experiments validate the effectiveness of our approach and the dataset's exceptional generalizability.
\end{abstract}

\vspace{-0.1in}

\vspace{-0.1in}
\section{Introduction}
\begin{figure}[!t]
\centering
\includegraphics[width=0.485\textwidth]{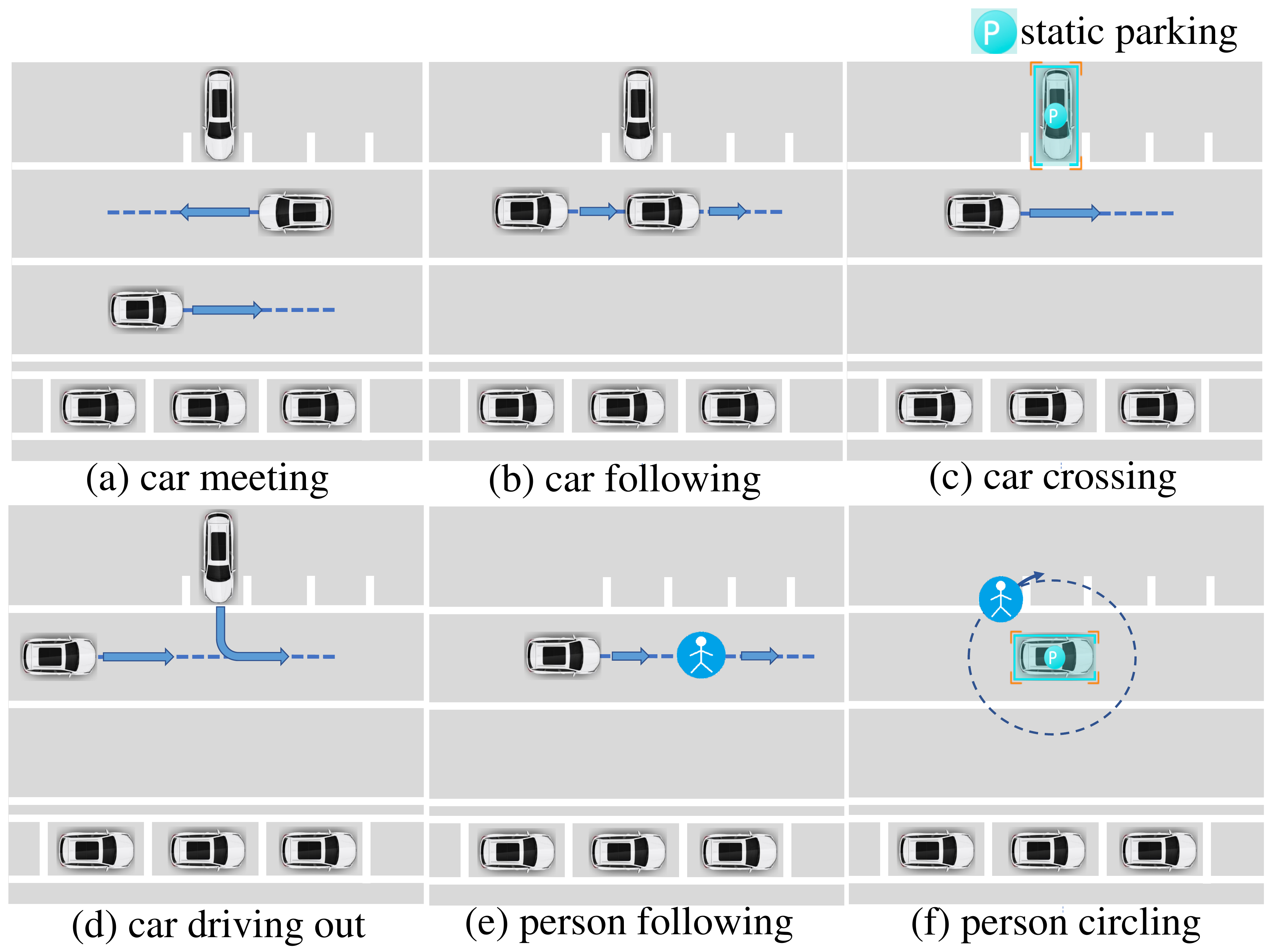}
\caption{Several valet parking scenes in the parking lots, including car meeting, car following, car crossing, car driving out, person following and person circling etc.}
\label{fig:scene}
\vspace{-0.3in}
\end{figure}

As the priority of developing an effective and safe advanced driver assistance system (ADAS)\yz{\cite{hoffmann2020real,wang2020soft,gao2022monocular}}, valet parking attracts more attention from industry and research communities in recent years \cite{song2019apollocar,zhou2020joint,wu2020motionnet}.
Among various driving assistance applications, valet parking exhibits an essential and challenging task.
Figure \ref{fig:scene} shows several challenging scenarios during valet parking\yz{\cite{samal2021task,chen2019pedestrian,yang2022predicting}}.
Besides, the environmental conditions in parking scenes, such as light and opacity, significantly increase the difficulty of robust environment perception \cite{wu2020psdet,chen2020collaborative}.
Unlike the relatively clear scenarios like highway and urban areas, valet parking aims to drive the vehicle into the drop-off area such as parking slots, which faces high requirements in perception \cite{kumar2021omnidet}.

Recent advances \cite{li2016vehicle, chen2016monocular} demonstrate the potential to replace the LiDAR with cheap onboard cameras, which are readily available on most modern vehicles\yz{\cite{su2018illumination,hehn2021fast}}.
Particularly, the surround-view fisheye cameras can provide a wider field-of-view (FoV) \cite{2018IVreal} than the pinhole cameras, which have grown popular in mass production.
In addition, four surround-view fisheye cameras cover a 360-degree perception, which makes up for pinhole cameras' near-field perception insufficiency, especially under the valet parking conditions \cite{kumar2022surround}. 
However, the fisheye lens usually perform obvious radial distortion, which leads to substantial appearance distortion \cite{kumar2022surround,wu2021disentangling}, complicating the surrounding's recognition.
In order to fully exploit the fisheye paradigm, more researchers begin to explore the fisheye surround-view perception, such as the position and pose information of the vehicles or pedestrians \cite{kumar2021omnidet,rashed2021generalized}.

Current datasets like KITTI \cite{geiger2012kitti}, Cityscapes \cite{article24} etc. mostly capture images with pinhole cameras, which comfortably achieve clear and distinguishable images under the urban, rural, or highway driving scenes. 
There are scarce fisheye datasets for autonomous driving \cite{yogamani2019woodscape,kitti360}, which have aided rapid progress for the fisheye surround-view perception.
Woodscape \cite{yogamani2019woodscape} and KITTI360 \cite{kitti360} collect large-scale fisheye datasets for different perception tasks on the ground.
However, these datasets place insufficient emphasis on perception with valet parking scenarios and fisheye image formation.
Hence, the models trained on public pinhole datasets or fisheye datasets on the ground may reveal suboptimal performance, since lacking sufficient training samples, particularly for valet parking scenes.

To expand surround-view fisheye perception tasks' images with various occlusions and postures under valet parking scenes, we contribute the first fisheye dataset called \textbf{F}isheye \textbf{P}arking \textbf{D}ataset (\textbf{FPD}) for parking scenarios, which has the following great properties, including (i) large scale quantity with more than 400 thousand fisheye images; (ii) high diversity with different parking lots, different periods and different parking conditions; (iii) high quality by filtering noisy and redundant images;
(iv) multiple kinds of annotations for different perception tasks, like 2D object detection, 3D object detection, BEV perception, depth estimation, etc.

Different from other public autonomous driving datasets \cite{geiger2012kitti,cordts2016cityscapes,caesar2020nuscenes}, our \textbf{FPD} dataset focuses on valet parking surround-view perception tasks and makes up the vacancy for the research in dealing with real-world parking lot scenes.
Furthermore, we offer the baseline on our \textbf{FPD} and propose the real-time distortion-insensitive multi-task network \textbf{F}isheye \textbf{P}erception \textbf{Net}work (\textbf{FPNet}), specifically for surround-view fisheye perception tasks, including 2D object detection, monocular 3D object detection, BEV perception, and monocular dense depth estimation.
The network achieves a balance between lightweight and accuracy, additionally with a particular module for addressing fisheye distortion.

Our contributions are summarized as follows:
\begin{itemize}
\item We build the first fisheye parking dataset \textbf{FPD}, concentrating on the surround-view fisheye perception including the 2D object detection, 3D object detection, BEV perception, and depth estimation.
Our contributed \textbf{FPD} comprises over 400 thousand fisheye images and contains attractive characteristics for parking scenarios.
\item We propose the baseline for our \textbf{FPD}: the distortion-insensitive multi-task framework \textbf{FPNet} for surround-view perception tasks, especially the BEV perception. 
\textbf{FPNet} utilizes the specific distortion module and lightweight designs to achieve real-time, distortion-insensitive and accurate performance.
\item  Comprehensive experiments validate the practicability of our collected \textbf{FPD} dataset and the effectiveness of \textbf{FPNet}.
\end{itemize}
\vspace{-0.1in}

\section{Related work}

\subsection{Autonomous driving datasets}
To meet the improving requirement for automated driving development, in the last decade, pioneer works create numerous datasets \cite{geiger2012kitti,caesar2020nuscenes,waymo,cordts2016cityscapes}, which cover the most autonomous driving tasks,
such as object detection \cite{geiger2012kitti, waymo}, semantic segmentation \cite{cordts2016cityscapes, caesar2020nuscenes}, depth estimation \cite{geiger2012kitti,nyuv2}, lane detection \cite{culane}, motion estimation \cite{caesar2020nuscenes, waymo} etc. , making amazing contribution to autonomous driving.

However, these datasets are almost pinhole camera datasets, with a limited field-of-view (FoV).  
Actually extensive visual tasks also adopt surround-view fisheye cameras to monitor the surrounding environment, due to their large FoV \cite{2018IVreal} and sufficiently stable performance \cite{2018IROSend}. 
In particular, ego-vehicles can achieve 360-degree perception using only four fisheye cameras, which is conducive to mass production \cite{2020ICRAfisheyedistancenet}. 
\yz{Moreover, the omnidirectional camera also captures the 360-degree field of view, which covers a full circle in the horizontal plane \cite{yang2020omnisupervised}.
}

\begin{table*}[t]
	\begin{center}
		\yz{\caption{The comparison of different imaging sensors.
		`FOV' denotes `field of view'. \\
		`H' denotes `horizontal FOV' and
		`V' denotes `vertical FOV'.	}\label{tab:lens_meterial}
		\vspace{-0.2in}
		{
	\begin{tabular}{c|c|c|c|c}
			\toprule
	         Name & FOV & Distortion &Dataset & Characteristics\\
			\midrule
		\multirow{2}*{pinhole camera} & \multirow{2}*{H:$<$180$^\circ$ V:$<$180$^\circ$} &\multirow{2}*{small} & KITTI\cite{geiger2012kitti} nuScenes \cite{caesar2020nuscenes} 
		&universal in most public perception methods;
		\\ ~ & ~ &~ 
		& Waymo \cite{waymo} Cityscapes \cite{cordts2016cityscapes} 
		& 
		caring little about the distortion      \\
        \toprule
        
		\multirow{2}*{fisheye camera} &\multirow{2}*{ H:$>$180$^\circ$ V:$<$180$^\circ$ }& \multirow{2}*{large} & \multirow{2}*{Woodscape\cite{yogamani2019woodscape} KITTI 360\cite{kitti360}} &common in surround-view system;  \\ ~ & ~ &~ & ~ &designing specific module to operate distortion   \\
		\toprule
		\multirow{2}*{omnidirectional camera} & \multirow{2}*{H:360$^\circ$ V:180$^\circ$}  & \multirow{2}*{large}  & Stanford2D3D\cite{Stanford2D3D} Matterport3D \cite{chang2017matterport3d}  &higher complexities and distortions implicated  \\ ~ & ~ &~ & 360D\cite{360D} PanoSUNCG\cite{PanoSUNCG} &in 360$^\circ$ full-view panoramas \cite{yang2020omnisupervised,stone2021deepfusenet}   \\
		\bottomrule
	\end{tabular}
	}}
	\end{center}
 	\vspace{-0.2in}
\end{table*}

Valeo releases the first fisheye dataset Woodscape \cite{yogamani2019woodscape}, to encourage the development of native fisheye models.
But Woodscape doesn't publish the LiDAR ground truth due to data protection restrictions.
KITTI360 \cite{kitti360} presents another large-scale dataset that contains rich sensory information including pinhole and fisheye cameras.
\yz{For omnidirectional cameras, the industry and academia provide the datasets like Stanford2D3D \cite{Stanford2D3D}, Matterport3D \cite{chang2017matterport3d}, 360D\cite{360D}, PanoSUNCG\cite{PanoSUNCG} etc. for the omnidirectional visual perception.}
However, these datasets generally focus on aboveground autonomous driving scenes, such as urban, rural, and highways.
There is no publicly available benchmark dataset for the valet parking scenes.
Our Fisheye Parking Dataset (\textbf{FPD}) could make up the vacancy, promoting research in dealing with real-world parking lot scenes.

\vspace{-0.1in}
\subsection{Monocular perception tasks} 
Autonomous driving involves various perception tasks, like object detection or depth estimation, to assist the system to cover a wider range of use cases.
In this paper, we mainly deal with three tasks: 2D object detection, monocular 3D detection, and monocular depth estimation.
We can derive the BEV perception from the 3D detection results.

\subsubsection{2D object detection}
2D object detection acts as a basic vision task.
CNN-based 2D object detection frameworks consist of one-stage detection methods \cite{article3,article33} and two-stage detection methods \cite{FastRCNN,ren2015faster,he2017mask}. 
As an end-to-end pipeline, one-stage methods achieve a significant trade-off between performance and speed, like SSD series \cite{article28,liu2016ssd,article49}, YOLO series \cite{article5,article33} and RetinaNet \cite{article3}.
In addition, two-stage methods, like RCNN series \cite{article7,article10,he2017mask}, take advantage of the predefined anchors to improve the performance at the cost of speed. 
Furthermore, \cite{article46,feng2022encoder} fuse multi-scale feature maps to improve detection with different scales.

\subsubsection{Monocular 3D detection}
Many prior works \cite{chen2016monocular,ku2019monocular,brazil2019m3d,simonelli2019disentangling} have tackled the inherently ill-posed problem of detecting 3D objects from monocular images.
Due to the lack of depth information from images, monocular 3D detection learns harder than LiDAR-based and stereo-based counterparts.
Many works \cite{liu2019deep,cai2020monocular,zhang2021objects} address this problem by utilizing 2D-3D geometric constraints to improve 3D detection performance.
In addition, CenterNet \cite{zhou2019objects} proposes a center-based anchor-free method but with restrained accuracy.
Following this work, center-based series SMOKE \cite{liu2020smoke}, KM3D \cite{li2021monocular} and RTM3D \cite{li2020rtm3d} assist the regression of object depth by solving a Perspective-n-Point method and have achieved remarkable results.
However, most existing works target the pinhole cameras instead of the fisheye ones, where fisheye cameras have a strong radial distortion and exhibit more complex projection geometry \cite{arsenali2019rotinvmtl}, which leads to appearance distortion \cite{plaut2020monocular}.

\subsubsection{Monocular depth estimation}
CNN-based supervised methods \cite{gu2021dro,lee2021patch,patil2022p3depth} are popular in monocular depth estimation tasks due to their superior performance. 
As a pioneer, Eigen et al. \cite{eigen2014depth} directly regress depth by employing two stacked deep networks for a coarse prediction, then refining it locally. 
Then Laina et al. \cite{laina2016deeper} adopt the end-to-end single CNN architecture, following the residual learning. 
Moreover, 
DRO \cite{gu2021dro} introduces a deep recurrent optimizer to alternately update the depth and camera poses through iterations, to minimize the feature-metric cost. 
Furthermore, the recent works \cite{ranftl2021vision,yang2021transformers} apply vision transformers to improve depth estimation.

\begin{figure}[!t]
\centering
\includegraphics[width=0.49\textwidth]{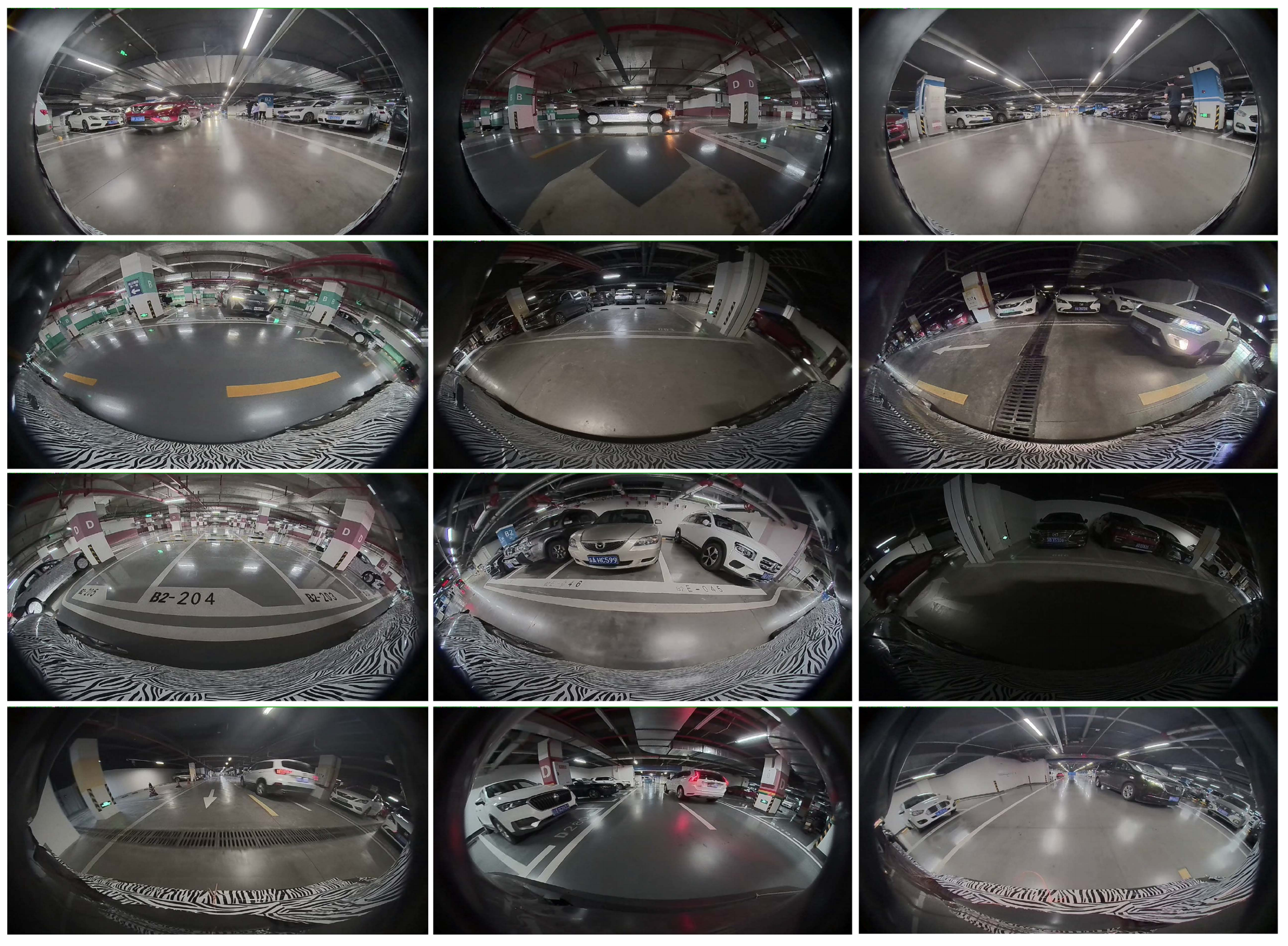}
\vspace{-0.2in}
\caption{Illustration of our collected real valet parking scenes, with different parking conditions, different lighting conditions, and different occlusions.}
\label{data_collection}
\vspace{-0.2in}
\end{figure}

\yz{
\subsection{Visual perception for different imaging sensors}
Most public perception methods design modules for the pinhole images (subsection B), which conduct the perspective projection and care little about the camera distortion. 
However, when utilizing wide-angle cameras (the fisheye camera or omnidirectional camera), researchers must concern about the large distortion and other issues.
The surround-view autonomous driving system usually adopts the fisheye cameras to achieve the surrounding perception \cite{2018ECCVWsemantic,2018CVPRWnear, 2021WACVsyndistnet,2020accessfisheyedet,2019arxivfisheyemodnet,2020CVPRWvehicle}, which builds specific components to operate the distortion \cite{dooley2015blind,yang2021efficient,mariotti2020spherical,kumar2021svdistnet}.
The omnidirectional camera reveals the higher complexities and distortions implicated in 360$^\circ$ full-view panoramas \cite{yang2020omnisupervised,stone2021deepfusenet}, where the omnidirectional images' perception also deals with the distortion and the problems caused from the large resolution, image preprocessing and knowledge transformation, etc.
\cite{lo2021efficient,li2020deep,360D,shere2021temporally,xu2021saliency}.
Table \ref{tab:lens_meterial} shows more comparison of the three imaging sensors.
}

\vspace{-0.1in}
\subsection{Multi-task visual perception}
Numerous autonomous driving investigations address complex real-world scenarios by joint learning of different subtasks \cite{mao2020multitask, vandenhende2021multi,teichmann2018multinet, kendall2018loss,sistu2019neurall,keqiang2022mjp}. 
MultiNet \cite{teichmann2018multinet} accomplishes road segmentation, detection, and classification tasks via a universal efficient architecture. 
NeurAll \cite{sistu2019neurall} brings unified CNN architecture including object recognition, motion,  depth estimation, and facilitating visual SLAM.
Li et al. \cite{keqiang2022mjp}, in like manner, propose a unified end-to-end framework (MJPNet) that shares predictions among multiple subtasks. 
These studies validate the benefit of joint multi-task learning in different autonomous driving scenes.

However, most studies focus on undistorted pinhole visual conditions and infrequent exploration in wide-angle fisheye camera-based perception.
OmniDet \cite{kumar2021omnidet} forms an encoder-shared framework with six primary tasks on the fisheye dataset WoodScape \cite{yogamani2019woodscape}.
On the whole, fisheye models reveal a significant growth potential with rich scenario distributions.

\begin{figure}[!t]
\centering
\includegraphics[width=0.495\textwidth]{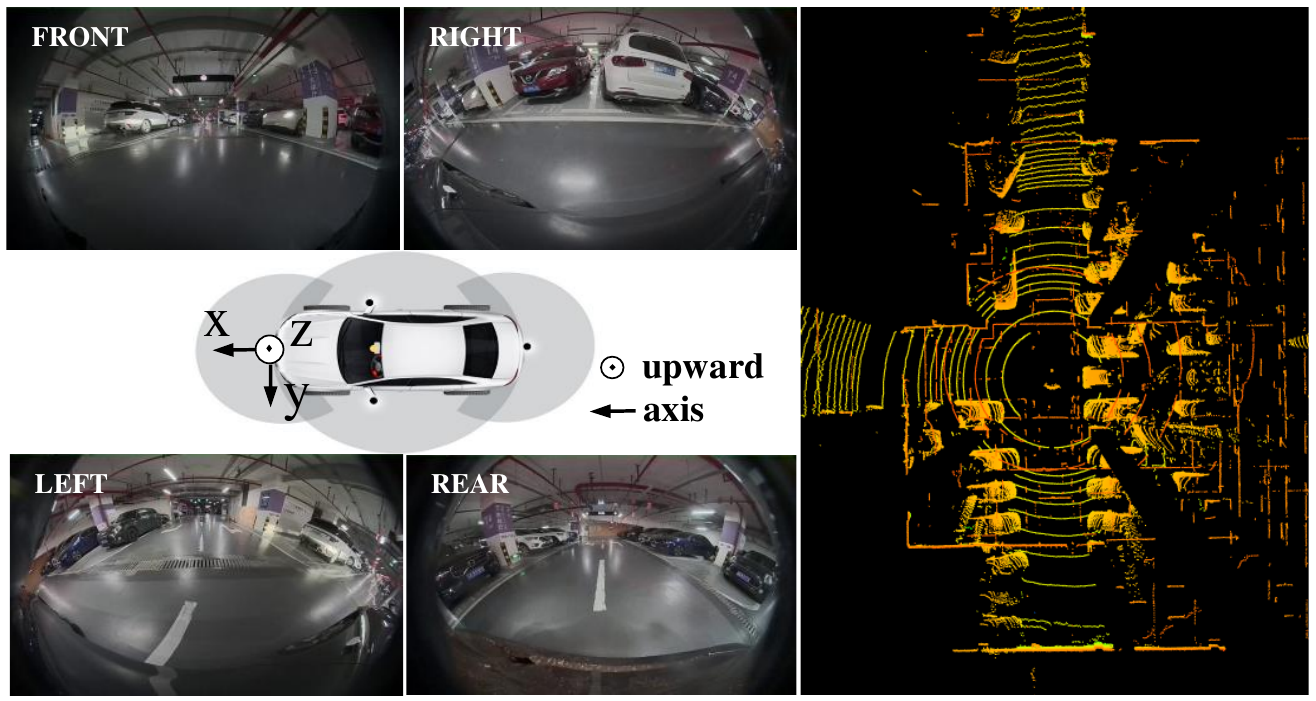}
\vspace{-0.1in}
\yz{\caption{Sensors' installation positions and produced images or point cloud's visualization, where the center of front bumper acts as the origin of the ego-vehicle coordinate system.}
}
\label{sensor}
\vspace{-0.1in}
\end{figure}

\section{Fisheye Parking Dataset}
In this section, we introduce our Fisheye Parking Dataset (\textbf{FPD}) in detail, including the process of data collection and annotation, dataset description and outstanding characteristics.

\vspace{-0.1in}
\subsection{Data Collection}

To ensure the diversity of autonomous driving scenarios, we collect a total of three cities, over one hundred parking lots, two periods (daytime and nighttime), and capture more than four hundred videos together with the point cloud sequences from LiDAR. 
Figure \ref{data_collection} displays several our collected real fisheye images of valet parking scenes.

Specifically, our master LiDAR adopts the RoboSense RS-Ruby, which has 128 beams, 10Hz capture frequency, $360^{\circ}$ horizontal FOV and $-25^{\circ}$ to $+15^{\circ}$ vertical FOV.
Besides, we choose four ZongMu fisheye RGB cameras, with 1920$\times$1280 resolution and 20Hz capture frequency.
Figure 3 demonstrates these sensors' installation positions and their produced images or point cloud's performance.
During the data recording process, the system aligns the timestamp between the videos from cameras and point cloud sequences from LiDAR, so qualified for the following annotation.
\yz{Furthermore, we conduct the calibration process of the sensors in the following three steps.
Firstly, we can directly calculate the camera intrinsics from the lens' distortion lookup table provided by the initial factory settings.
Secondly, we achieve the extrinsics (x, y, z, pitch, yaw, roll) of the LiDAR and cameras based on the ego-vehicle coordinate system via the measurement equipment.
Thirdly, we further rectify the camera's extrinsics by the alignment of projected LiDAR points and images' semantics.
As shown in Figure 5, we project the LiDAR points to the image plane via the above calibrations, then we manually adjust the camera's extrinsics to match the projected points and the semantic contents.

}
\begin{figure*}[!t]
\centering
\includegraphics[width=0.99\textwidth]{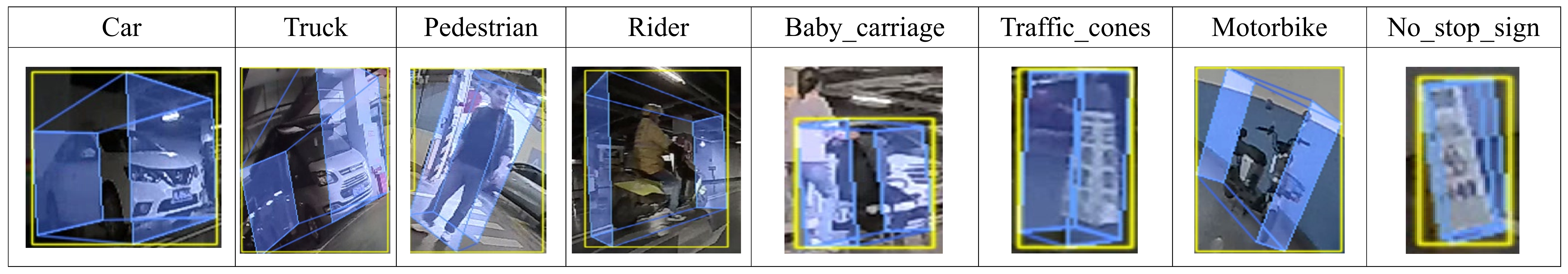}
\vspace{-0.1in}
\caption{Our data annotation contains eight categories, including the car, truck, pedestrian, rider, baby carriage, traffic cones, motorbike and no-stop sign.}
\label{fig:categories}
\vspace{-0.1in}
\end{figure*}
\begin{figure}[!t]
\label{calibration_3}
\centering
\vspace{-0.1in}
\includegraphics[width=0.46\textwidth]{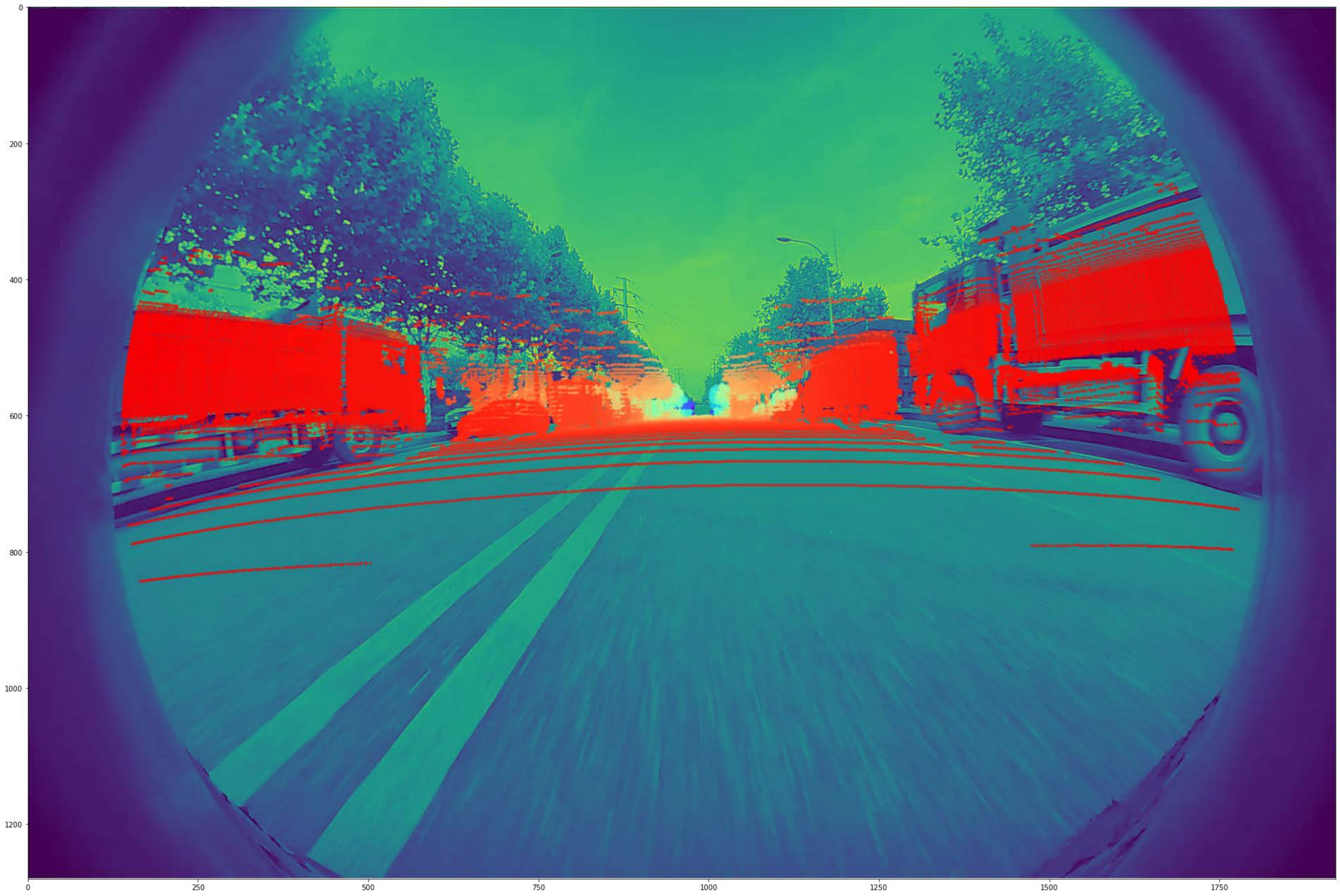}
\yz{
\caption{We project the LiDAR points to the camera image plane to rectify the camera's extrinsics where the color denotes the points' different distances.}
}
\vspace{-0.1in}
\end{figure}
\begin{figure}[!t]
\centering
\includegraphics[width=0.49\textwidth]{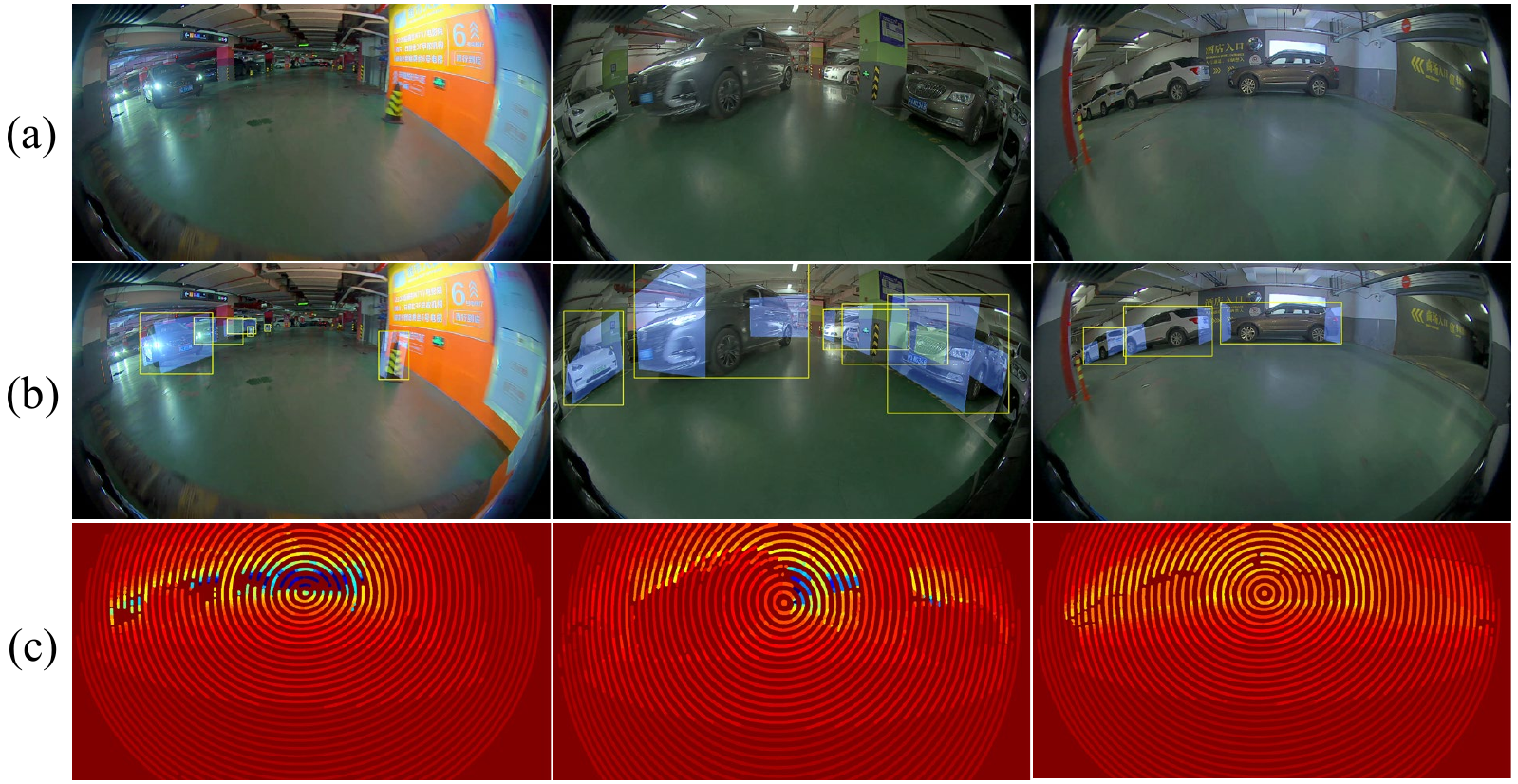}
\caption{Visualization of our annotated labels. (a) our captured monocular fisheye images; (b) associated 2D object labels (yellow bbox) and projected 3D object labels (blue bbox); (c) associated depth ground truth after depth completion.}
\label{fig:objects_annoation}
\vspace{-0.2in}
\end{figure}

To cover various realistic scenes of the parking lot, we artificially arrange kinds of driving scenes to collect data, like car meeting, car crossing and person circling, etc., as shown in Figure \ref{fig:scene}, which are common but critical cases for autonomous parking task.

\vspace{-0.3in}
\subsection{Data Annotation}
We annotate the dataset in the same way as KITTI dataset \cite{geiger2012kitti}, by drawing a tight bounding box around each object's complete point cloud body.
We don't cover all objects' continuous moving process for redundant annotations.
Instead, we remove similar segments and annotate data with intervals of three to five frames.
In addition, we restrict the visible range (within 15 meters), so we abandon too far-away objects.
For occluded objects, we reserve the 3D bounding box if the occlusion rate performs less than 80\%, by way of imagining the full 3D bounding box with annotators' experience.
Figure \ref{fig:objects_annoation} shows several examples of our annotated labels.

Our annotation contains eight categories, including the car, truck, pedestrian, rider, baby carriage, traffic cones, motorbike and no-stop sign.
Figure \ref{fig:categories} demonstrates annotation demos of the eight categories, where the blue bounding boxes mean the 3D annotation's 2D visualization, projected from point cloud plane to image plane.
Yellow bounding boxes indicate the outer bounding rectangle of blue projected points, as our 2D object detection ground truth, also shown in Figure \ref{fig:objects_annoation} (b).

\begin{figure}[!t]
\centering
\includegraphics[width=0.4\textwidth]{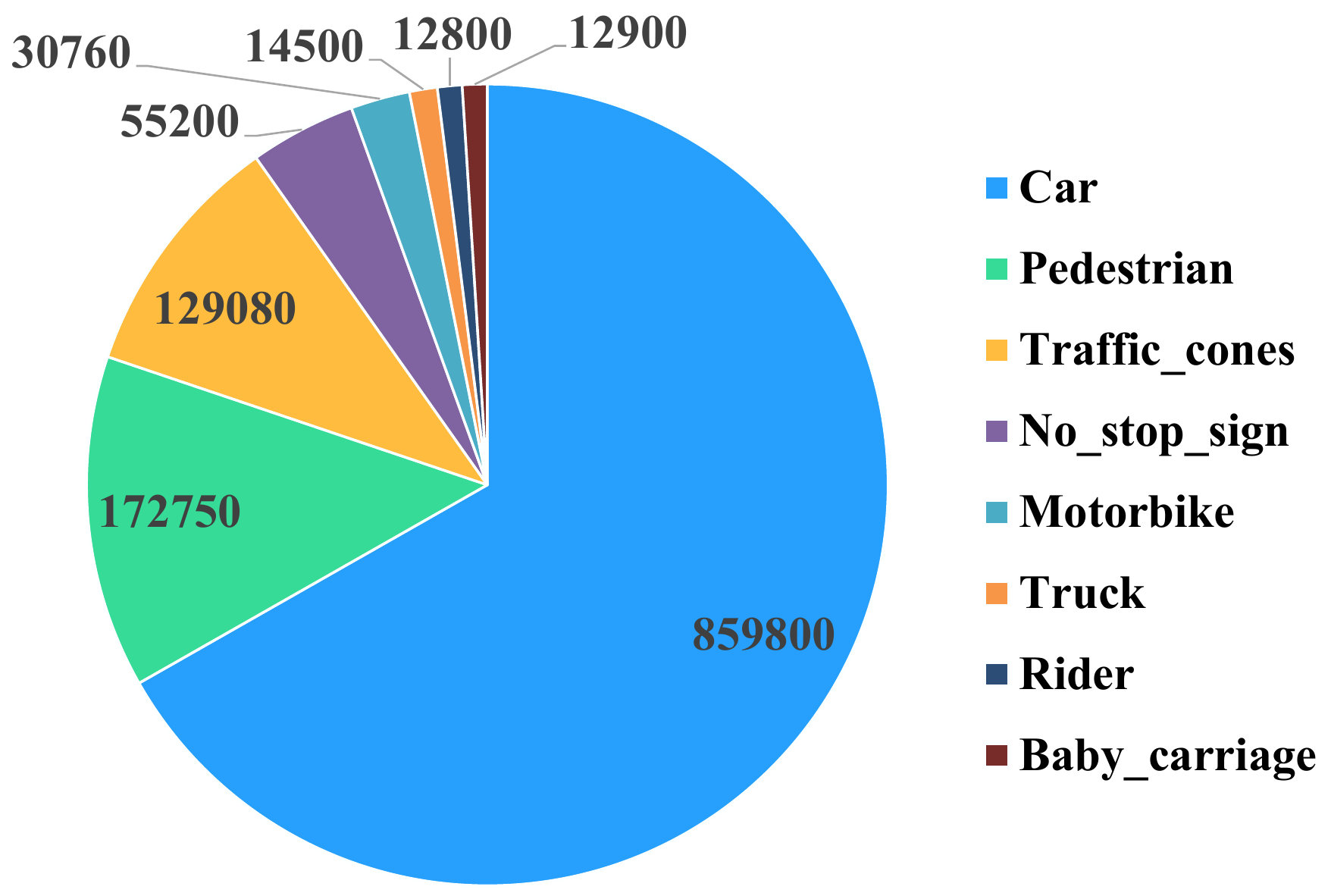}
\vspace{-0.1in}
\caption{The detailed data distribution of \textbf{FPD} in different categories.}
\label{fig:objects_summary}
\vspace{-0.1in}
\end{figure}

\begin{table}[t]
	\begin{center}
		\caption{Statistics of data annotations with our \textbf{FPD}. We obtain more than 400,000 annotations totally.}\label{tab:complexity-ana}
		{
		\begin{tabular}{c|ccc|c}
			\toprule
	         & Training & Validation & Testing & Sum\\
			\midrule
			\# images & 210,000 & 126,000 & 84,000 & 420,000 \\

			\bottomrule
	\end{tabular}
	}
	\end{center}
	\vspace{-0.3in}
\end{table}

Furthermore, we project the point cloud to the monocular images with the assistance of calibration and distortion parameters for the sparse depth map.
Then we adopt the depth completion method IP-Basic \cite{ku2018defense} to create more robust depth ground truth, as shown in Figure \ref{fig:objects_annoation} (c).

\vspace{-0.1in}
\subsection{Dataset Description}

Table \ref{tab:complexity-ana} and Figure \ref{fig:objects_summary} illustrate the statistics of our \textbf{FPD}. 
Totally, we obtain more than 400,000 data, where one data contains four fisheye images and one point cloud with annotation.
In addition, one data accompanies one intrinsic parameter, one extrinsic parameter and one fisheye distortion parameter.
We can project the point cloud annotation to the image via the intrinsic, extrinsic and fisheye distortion parameters to obtain the 2D object bounding box and depth ground truth.

Moreover, we split the \textbf{FPD} into training, validation and testing sets with the ratio of 5:3:2 and the amount of 210,000, 126,000, and 84,000.
The ratio of daytime and nighttime scenes reveals 2:1.
In addition, an average of 4 thousand data per parking lot composes more than 400,000 annotations, where the most frequent categories indicate the cars, the pedestrians, and the traffic cones, as shown in Figure \ref{fig:objects_summary}. 
\begin{figure*}
\centering
\includegraphics[width=0.98\textwidth]{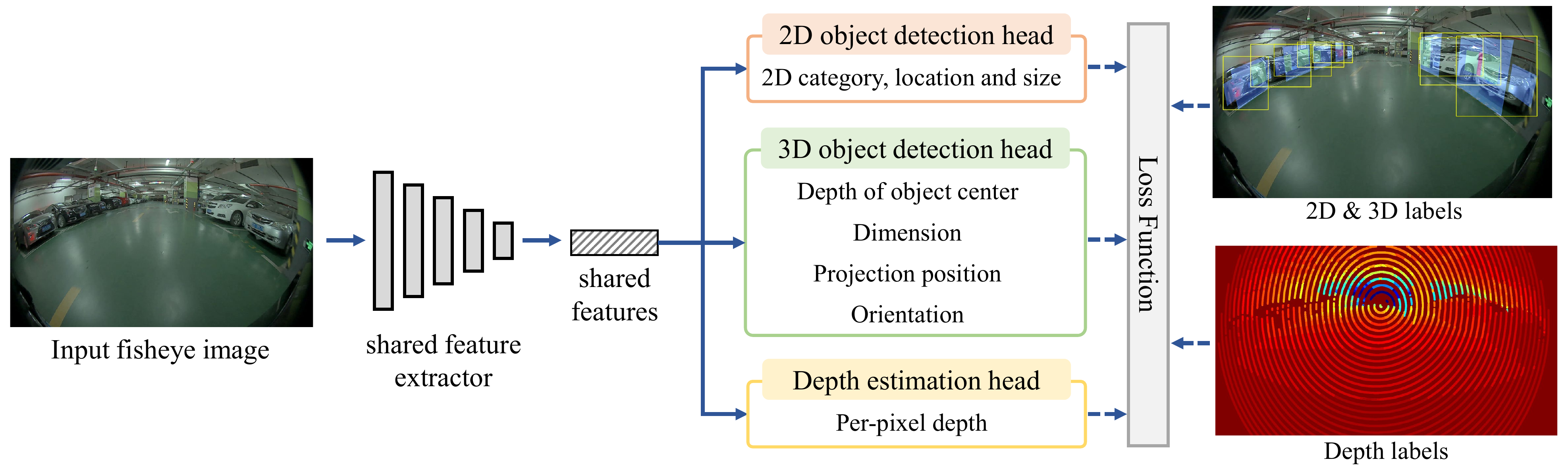}
\vspace{-0.1in}
\caption{
The framework of our FPNet, which mainly consists of a shared feature extractor and a multi-task perception module, including a 2D object detection head, a 3D object detection head and a depth estimation head.
}
\label{fig:distribution}
\vspace{-0.2in}
\end{figure*}

\vspace{-0.1in}
\subsection{Dataset Characteristics}
As the first large-scale real-world fisheye dataset, our compiled \textbf{FPD} exhibits the following nice properties:
\subsubsection{First fisheye dataset for parking scenarios}
We provide the first fisheye dataset \textbf{FPD}, 
which focuses on multiple autonomous driving tasks in parking scenarios, also distinct from natural scenes of public datasets.
Environmental conditions in parking scenarios, such as light and opacity, significantly increase the detection difficulty.
Concerning a variety of tough parking scenarios, \textbf{FPD} can promote research in dealing with real-world parking problems.

\subsubsection{Great quantity}
So far, our \textbf{FPD} contains more than 400 thousand data from over 200 hours of parking scene videos and point cloud sequences.
In the future, we will continue to collect diverse parking to enrich the existing dataset.

\subsubsection{High quality and diversity}
Our \textbf{FPD} covers three cities, over one hundred parking lots from different periods, and different parking cases. 
Besides, we carefully pick out high-quality images and point clouds with high resolution, ensuring our dataset's superiority.

\subsubsection{Multi-purpose}
As a point-cloud-based dataset, our \textbf{FPD}'s potential not only lies in the three tasks (i.e., 2D object detection, monocular 3D object detection and depth estimation) but also in other vision tasks, such as point cloud 3D object detection, 2D or 3D semantic segmentation, video object detection. 
Therefore, the \textbf{FPD} is multi-purpose for diverse tasks.

\vspace{-0.1in}
\section{Distortion-insensitive
Multi-task Framework}
In this section, we introduce our surround-view fisheye monocular distortion-insensitive multi-task framework FPNet. 
\vspace{-0.1in}


\vspace{-0.2in}
\subsection{Overview}
Figure \ref{fig:distribution} demonstrates the framework of our FPNet, which mainly consists of a shared feature extractor and a multi-task perception module, including a 2D object detection head, a 3D object detection head and a depth estimation head.


Given a fisheye monocular image $I \in R^{3\times H_I\times W_I}$, we directly adopt our shared feature extractor to obtain features ${F^{2d}\in R^{C\times H_{F}\times W_{F}}}$, prepared for the following 2D object detection, 3D monocular object detection (BEV perception) and dense depth estimation.
Through the multi-task perception head, we complete the predictions of three tasks (Section \ref{network}).
During training, we project 3D ground truth to the monocular image plane to create the predictions' supervision, where the fisheye distortion module operates the distortion of fisheye camera projection. 
During testing, we utilize the post-processing module to decode the network's prediction, together with the fisheye distortion module's operation (Section \ref{training and testing} and Section \ref{fisheye_dis}).
Furthermore, we deploy our model to the embedded system with Qualcomm 8155 chip, to achieve real-time and excellent perception performance (Section \ref{deployment}).

\vspace{-0.1in}
\subsection{Network}
\label{network}

\subsubsection{Shared feature extractor}
To balance the trade-off between performance and speed, we select DLA34 \cite{yu2018deep} as our shared feature extractor. 
Additionally, we apply some improvements to achieve the lightweight requirement.
Firstly, we downsample the input image with radio 8 instead of the usual 4, which saves lots of time but maintains the accuracy (see Section \ref{exp_share}).
Secondly, we get rid of some redundant layers according to their inference time on the embedded device, still with great performance (see Section \ref{exp_share}).


\subsubsection{Multi-task perception head}
After extracting features from fisheye monocular images, we make predictions for our three perception tasks via multi-task perception head.


\textbf{2D object detection and monocular 3D object detection.}
2D object detection mainly focuses on searching for objects' bounding box in the image, while monocular 3D object detection tries to locate objects' 3D position, and regress objects' dimension and orientation, which obviously performs more difficult compared to 2D object detection.
For shared feature extractor design and real-time requirements, the center-based framework exhibits more comfortable for our 2D and 3D object detection.
Specifically, inspired by MonoCon \cite{liu2021learning}, we build our 3D object detection head, whose input remains the shared features. 
Different from MonoCon \cite{liu2021learning}, we directly predict the projected 3D bbox center heatmap, instead of the 2D bbox center heatmap and offset from projected 3D center to 2D center, which enhances the detection accuracy (see Section \ref{exp_bbox}).
Then we reserve the 3D-related predictions of object depth and uncertainty, shape dimensions and heading angle (MultiBin \cite{mousavian20173d} regression).
For 2D object detection, to reduce computation, we predict the offset from 2D center to the projected 3D center, then we can obtain the 2D center prediction by adding the above projected 3D center prediction and this offset.
Besides, we also predict 2D bbox's height and width.
At last, we abandon other auxiliary monocular contexts \cite{liu2021learning}, also remaining satisfying performance (see Section \ref{exp_share}).

\textbf{Monocular depth estimation.}
Different from the above object centers' depth estimation, we also estimate the dense depth in this task.
We refer DRO \cite{gu2021dro} to our depth estimation model, which indicates a gated recurrent network and iteratively updates the depth map between two images by minimizing a feature-metric cost \cite{gu2021dro}.
So we are essential to cache the front frame image, to meet the tuple input of the depth estimation model.
In addition, we replace the original feature backbone with our shared feature extractor, to fulfill multi-task architecture.
Furthermore, the depth estimation model also suits the self-supervised task. 
But in this work, we focus on supervised depth estimation.

\vspace{-0.1in}
\subsection{Training and testing}
\label{training and testing}
\subsubsection{Training and loss function}
For 2D object detection and 3D object detection, we settle on different loss function for different subtasks as follows:

($\romannumeral1$) The focal loss for projected 3D bbox center heatmap:
\begin{equation}
\mathcal{L}(\mathcal{H},\mathcal{H}^*) = \frac{-1}{N} \sum_{(x,y)}\begin{cases}
(1-\mathcal{H}_{xy})^\gamma\log(\mathcal{H}_{xy}),\ if\ \mathcal{H}_{xy}^* = 1,\\
(1-\mathcal{H}_{xy}^*)^\beta(\mathcal{H}_{xy})^\gamma\log(1-\mathcal{H}_{xy}), else\\
\end{cases}
\label{equ_1}
\end{equation}

To acquire the ground truth of projected 3D bbox center, we project the 3D bbox labels to the image plane via the fisheye distortion module.
Then we follow CenterNet \cite{duan2019centernet} to generate the ground-truth heatmap $\mathcal{H}^*\in R^{1\times H_{F}\times W_{F}}$.
$\mathcal{H}\in R^{1\times H_{F}\times W_{F}}$ means the predicted heatmap.
We adopt the focal loss (Equ. \ref{equ_1}) where the $\alpha$ and $\beta$ are hyper-parameters ($\alpha$ = 4.0 and $\beta$ = 2.0).
We will detail the fisheye distortion module in section \ref{fisheye_dis}.

($\romannumeral2$) The L1 loss for the center offset, 2D bounding box's width and height, 3D bounding box's width, height and length, the intra-bin angle residual in heading angles, the dense depth estimation:
\begin{equation}
\mathcal{L}(\mathcal{S},\mathcal{S}^{*})=\lambda\cdot\|S-S^{*}\|_1,
\end{equation}
where $\mathcal{S}^{*}$ indicates subtasks' ground truth and $\mathcal{S}$ means the predicted values.

($\romannumeral3$) The Laplacian aleatoric uncertainty loss function for the object depth estimation.
Following \cite{liu2021learning}, we use the Laplace distribution to model the uncertainty and optimize the depth and uncertainty at the same time.
 

($\romannumeral4$) The cross-entropy loss function for the bin index in heading angles.
We assign the bin index task as a classification task, so we adopt the cross-entropy loss function.

\subsubsection{Testing and post-processing}
During testing, except for the dense depth estimation directly predicted from the network, we ought to utilize the post-processing to produce final 2D and 3D detection results.
Specifically, we first infer the network to output the projected 3D center's heatmap.
Then we calculate the local maximum of the heatmap and select the top-k's positions as the predicted 3D centers.
According to 3D centers' positions, we can obtain from the predictions: the offset from the 2D center to the projected 3D center; 2D bbox's width and height; 3D bbox's width, height, and length; 3D object's depth, uncertainty, heading angles' bin index and intra-bin angle residual. 
Through the fisheye distortion module, the projected 3D center, together with the predicted object depth, can resume the 3D position. 
In addition, the object confidence depends on both the projected 3D center's score in the heatmap and the uncertainty, as follows:
\begin{equation}
    C_{obj} = C_{proj\_center}* e^{(-\sigma)}
\end{equation}
where $\sigma$ denotes the uncertainty.
For heading angles, we recover the rotation with the predicted bin index and intra-bin residual, referring to MultiBin \cite{mousavian20173d}.
Furthermore, we project the 3D results to the Bird's-Eye-View (BEV) plane to obtain the BEV perception. 
Actually, when given the front, left, right and rear fisheye images at the same time, we can generate the 360-degree BEV perception after some fusion procedures like ReID or object filtering.

\vspace{-0.1in}
\subsection{Fisheye distortion module}
\label{fisheye_dis}
We achieve the distortion-insensitive function by our \textbf{F}isheye \textbf{D}istortion \textbf{M}odule (\textbf{FDM}), which correctly builds fisheye projection between 3D space and 2D image plane to exclude the distortion's disturbing.
This module has two main functions: (i) producing 2D labels from projecting 3D ground truth; (ii) restoring 3D positions from 2D image points. 
Compared with pinhole model projection, fisheye model projection needs to consider the influence of fisheye distortion.
To achieve these purposes, we summarize our procedures of fisheye projection.

\textbf{Task (i)}. Given the 3D point $(x,y,z)$ in the camera coordinate system and camera parameters, we want to solve out the 2D point $(u,v)$ in the image plane. 
According to the \cite{kumar2022surround}, for pinhole projection, we demand to calculate the field-angle $\theta$ of the imaged point (the field angle of the projected ray \cite{kumar2022surround}):
\begin{equation}
    \theta = a\tan(r)
\end{equation}
The pinhole projection coordinates $(a, b)$ denote $a=x/z$, $b=y/z$ and $r=\sqrt{a^2+b^2}$.
Then for fisheye distortion projection, we utilize the fisheye distortion parameters to compute the rectified angle $\theta_d$:
\begin{equation}
    \theta_d = \theta (1 + k_1 \theta^2 + k_2 \theta^4 + k_3 \theta^6 + k_4 \theta^8)
\end{equation}
Afterwards, we obtain the distorted point coordinates $(x^{\prime}, y^{\prime})$:
\begin{align}
    x^{\prime} = (\theta_d/r)a \\
    y^{\prime} = (\theta_d/r)b 
\end{align}
Finally, we achieve the final pixel coordinates vector $(u,v)$:
\begin{equation}
    \begin{bmatrix}
    u\\
    v\\
    1\\
    \end{bmatrix}
    =
    \begin{bmatrix}
    f_u & 0 & c_u \\
    0 & f_v & c_v \\
    0 & 0 & 1 \\
    \end{bmatrix}
    \begin{bmatrix}
    x^{\prime}\\
    y^{\prime}\\
    1\\
    \end{bmatrix}
\end{equation}
where $f_u, f_v, c_u, c_v$ are camera's intrinsic parameters.
So according to Equ.(4)-(8), we successfully construct the projected 2D ground-truth coordinates $(u,v)$ based on distorted ground-truth 3D coordinates $(x,y,z)$.
\begin{figure}[!t]
\centering
\includegraphics[width=0.48\textwidth]{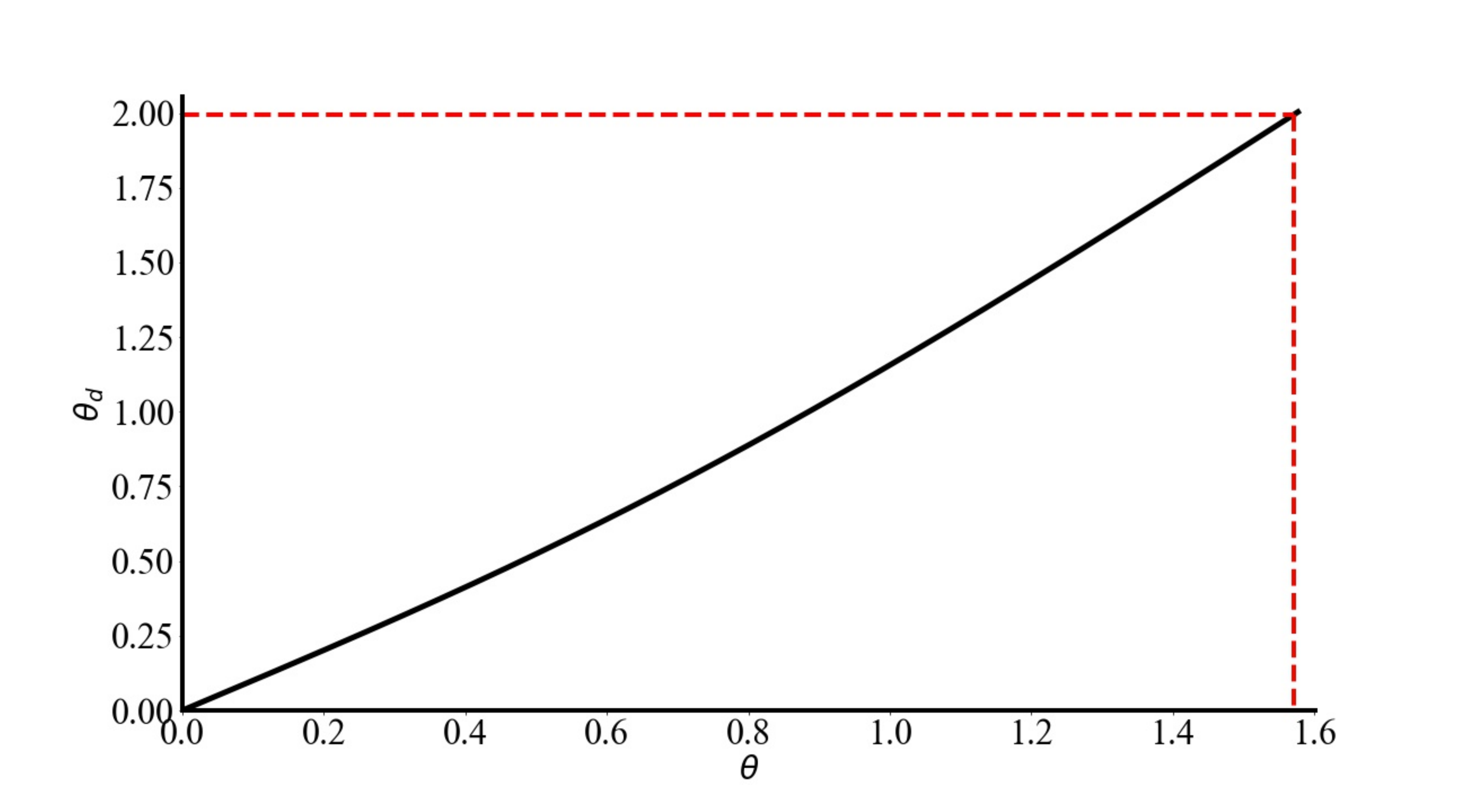}
\vspace{-0.1in}
\caption{The transformation between $\theta$ and $\theta_d$: $\theta_d$ is monotonically increasing with $\theta$, where $\theta$ ranges from 0 to $\pi/2$. }
\label{lut}
\vspace{-0.2in}
\end{figure}

\textbf{Task (ii)}. Given the 2D image point $(u, v)$, point's depth $z$ and camera parameters, we need to recovery the 3D position $(x, y, z)$.
Firstly, we compute the distorted point coordinates $(x^{\prime}, y^{\prime})$, according to the inverse operation:
\begin{equation}
    \begin{bmatrix}
    x'\\
    y'\\
    1\\
    \end{bmatrix}
    =
    \begin{bmatrix}
    f_u & 0 & c_u \\
    0 & f_v & c_v \\
    0 & 0 & 1 \\
    \end{bmatrix}^{-1}
    \begin{bmatrix}
    u\\
    v\\
    1\\
    \end{bmatrix}
\end{equation}
Then combining the Equ.(6) and (7), we can acquire the $\theta_d$, as follows:
\begin{align}
(x')^2+(y')^2 =(\theta_d)^2(a^2+b^2)/r^2= (\theta_d)^2 \\
\theta_d = \sqrt{(x')^2+(y')^2}
\end{align}
So in Equ. (5), we have to solve out the $\theta$ with the given $\theta_d$, with a mathematical tool's help.
Additionally, we restrict the solution $\theta$ under the filter of the unique value ranging from 0 to $\pi/2$ in reality, and finally we achieve the unique value $\theta$.
Afterwards, we can calculate $a$ and $b$ from two equations:
\begin{align}
    a = (x^{\prime} \cdot r)/\theta_d  \\
    b = (y^{\prime} \cdot r)/\theta_d 
\end{align}
where $r=\sqrt{a^2+b^2}$.

At last, we acquire the 3D position $(x,y,z)$ by:
\begin{align}
    x=a \cdot z \\
    y=b \cdot z
\end{align}

So according to Equ. (9)-(15), we complete the precess of our task (ii) for the fisheye distortion module.

\textbf{Speeding up}.
Reviewing the above procedures, we detect that it is time-consuming for Equ. (5), especially when solving out the $\theta$ given $\theta_d$.
However, on closer observation (see Figure \ref{lut}), $\theta_d$ is monotonically increasing with $\theta$, where $\theta$ ranges from 0 to $\pi/2$. 

So for speeding up our framework, we further introduce a look-up table indexing approach. Specifically, we build a look-up table consisting of 900 grids in which $\theta_d$ can be found with corresponded $\theta$ or $\theta$ with corresponded $\theta_d$.
The procedure of looking up is economical of time, enhancing the real-time performance (see Section \ref{exp_share}).

\vspace{-0.1in}
\subsection{Deployment}
\label{deployment}
We deploy our model to the embedded system with Qualcomm 8155 chip, which only accepts the Qualcomm Deep Learning Container (DLC) file.
So we transform our PyTorch model to the ONNX model, then convert the ONNX model to the DLC file with the help of Qualcomm conversion tools.
In addition, we run our post-processing on the chip instead of the embedded system's CPU, to save CPU load (from 30ms to 3ms).
Furthermore, we artificially quantize the DLC file to satisfy 8155 chip's AIP mode and reduce inference errors compared with the automatically quantized DSP mode.

\vspace{-0.1in}
\section{Experiments}
This section exhibits our fisheye multi-task network FPNet on our \textbf{FPD}. 
Then we discuss \textbf{FPD}'s generalization across datasets.
Furthermore, we analyze the effects of lightweight design and fisheye distortion module FDM by ablation study and comparison.
Finally, we reveal several qualitative perception results by our approach.
\vspace{-0.1in}
\subsection{Evaluation Metrics}
For 2D object detection, we select the IoU criterion of 0.7 for object detection metrics: Average Precision (AP) and Average Recall (AR), indicating $AP_{2D}$ and $AR_{2D}$.
For 3D object detection, the 3D Average Precision ($AP_{3D}$) and BEV Average Precision ($AP_{BEV}$) are two vital evaluation metrics.
We utilize the IoU threshold of 0.5 for AP$_{40}$ with all categories. 
We adopt the depth metric for dense depth estimation, absolute relative error (abs rel.). Specially, less absolute relative error (abs rel.) means better performance, which acts different from the $AP_{2D}$, $AR_{2D}$, $AP_{3D}$ and $AP_{BEV}$.
We evaluate the hardware platform's inference speed with time consumption (ms).

\vspace{-0.1in}
\subsection{Implementation Details}
We conduct experiments with framework Pytorch on the Ubuntu system and employ eight NVIDIA RTX A6000s. 
The initial learning rate is set to be 0.1, and the momentum and learning decay rates are 0.9 and 0.01. 
We adopt stochastic gradient descent (SGD) solver and 16 batch size settings.
For preprocessing of initial images (width: 1920 pixels and height: 1280 pixels), we firstly crop the height (upper 200 pixels and lower 210 pixels) to remove the blank object space, then resize them to the final size: 640 pixels width and 480 pixels height. 

For 2D object detection, baseline detectors contain CenterNet \cite{duan2019centernet} and RetinaNet \cite{article3}. 
For 3D object detection, we choose KM3D \cite{li2021monocular} and MonoCon \cite{liu2021learning} as our baseline detectors.
For dense depth estimation, we pick DRO \cite{gu2021dro} as the baseline.
All the baselines have occupied the related field in recent years.
To ensure comparability, all baselines utilize the same experimental settings as their release.
Additionally, we insert our fisheye distortion module to make them suitable for fisheye images.
\begin{table}[!t]
\small
\caption{The performance of baselines on our FPD dataset with three visual tasks: 2D object detection, monocular 3D object detection, and monocular dense depth estimation. Highest result is marked with \textcolor{red}{red} and the second highest is marked with \textcolor{blue}{blue}. Different from the $AP_{2D}$, $AR_{2D}$, $AP_{3D}$ and $AP_{BEV}$, less abs rel. means better performance.}
\label{tab:baselines_result}
\centering

\begin{tabular}{c|ccccc}
\toprule
Method & $AP_{2D}$     & $AR_{2D}$     & $AP_{3D}$               & $AP_{BEV}$              & abs rel.               \\ \midrule
RetinaNet \cite{article3}           & \textcolor{blue}{49.3}  & 44.6  & - & -  & - \\
CenterNet \cite{duan2019centernet}  & 45.4  & 53.7  & - & -  & -    \\
KM3D \cite{li2021monocular}         & 46.0 & \textcolor{blue}{55.1}  & 58.25 & 73.41  & -  \\
MonoCon \cite{liu2021learning}      & 48.8 & 54.3  & \textcolor{blue}{63.71} & \textcolor{blue}{76.49}  & -  \\
DRO \cite{gu2021dro}               & -     & -  & - & -  & \textcolor{blue}{0.095}     \\
\midrule 
FPNet        & \textcolor{red}{\textbf{53.4}} & \textcolor{red}{\textbf{57.2}} & \textcolor{red}{\textbf{66.38}} & \textcolor{red}{\textbf{80.74}}  & \textcolor{red}{\textbf{0.088}} \\
Improvement                      & +4.1    &+2.1  &+2.67  &+4.25   &+0.007  \\
\bottomrule
\end{tabular}
\vspace{-0.1in}
\end{table}
\begin{table}[!t]
\small
\caption{Evaluation of the generalizability of the FPD dataset to public datasets based on CenterNet \cite{duan2019centernet} and DRO \cite{gu2021dro}.
A$\Rightarrow$B means adopting the model pre-trained on dataset A to finetune on dataset B.
}
\label{cross-dataset}
\centering
\begin{tabular}{c|ccccccc}
\toprule
Dataset
 & $AP_{2D}$     & $AR_{2D}$     & $AP_{3D}$               & $AP_{BEV}$              & abs rel.  \\
\midrule
 COCO \cite{lin2014microsoft} &45.2  &39.5 &- &- &-  \\ 
 FPD$\Rightarrow$COCO &\textbf{47.8}  &\textbf{43.2} &- &- &-     \\
\midrule
 KITTI \cite{geiger2012kitti}&80.6 &82.2 &48.01  &53.39  &0.056\\ 
 FPD$\Rightarrow$KITTI &\textbf{84.1} &\textbf{84.7} &\textbf{51.25} &\textbf{56.13} &\textbf{0.053} \\ 
\bottomrule
\end{tabular}
\vspace{-0.2in}
\end{table}
\begin{table}[t]
\small
\centering
\caption{Ablation experiments with two settings for the shared feature extractor: down-sampling radio (DR) and layer removal (LR).}
\begin{tabular}{c|cc|cccc}
    \toprule
    &DR  &LR  & $AP_{2D}$ & $AP_{3D}$   & abs rel. & $\text{time}$\\
    \midrule
    (a)& 4    &- & 53.7 & 66.47 & 0.076 &72\\
    (b)& 4  & \checkmark & 53.7& 66.46 & 0.076 &64 \\
    (c)& 8    &- & 53.5 & 66.40 & 0.078  &40\\
    (d)& 8    & \checkmark  & \textbf{53.4} & \textbf{66.38} & \textbf{0.078} & \textbf{23}\\
  \bottomrule
\end{tabular}
\label{lightweight1} 
\vspace{-0.1in}
\end{table}
\begin{table}[t]
\small
\centering
\caption{Ablation experiments with two settings for auxiliary monocular contexts: eight projected keypoints heatmap and offsets vectors estimation (8 Key.); and quantization residual estimation (Quan.).}
\begin{tabular}{c|cc|cccc}
    \toprule
    &8 Key. & Quan. & $AP_{2D}$ & $AP_{3D}$   & abs rel. & $\text{time}$\\
    \midrule
    (a)& \checkmark  & \checkmark & 53.6 & 66.50 & 0.079 &40 \\
    (b)& -           & \checkmark  & 53.5 & 66.41 & 0.078 &25  \\
    (c)& \checkmark    &  -      & 53.6 & 66.43 & 0.079 &34  \\
    (d)& - & -   & \textbf{53.4} & \textbf{66.38} & \textbf{0.078} & \textbf{23} \\
  \bottomrule
\end{tabular}
\label{lightweight2}  
\vspace{-0.2in}
\end{table}

Moreover, we choose two public datasets for our cross-dataset evaluation: COCO \cite{lin2014microsoft} and KITTI \cite{geiger2012kitti}.
COCO \cite{lin2014microsoft} dataset only contains 2D object detection task, while KITTI \cite{geiger2012kitti} covers the 3D object detection, 2D object detecion and depth estimation. 

\vspace{-0.1in}
\subsection{Results and Analysis}
\subsubsection{\textbf{Results on FPD}}
To demonstrate the effectiveness of our FPNet method, we conduct three tasks' baseline evaluation based on \textbf{FPD}, as shown in Table \ref{tab:baselines_result}.

\begin{table}[t]
\small
\centering
\caption{Comparison between our look-up table indexing and SolvePoly function from Opencv library \cite{bradski2000opencv} with FPNet on \textbf{FPD}. `LTI' indicates the `Look-up Table Indexing'.}
\begin{tabular}{c|c|cccc}
    \toprule
	 & Method & $AP_{2D}$ & $AP_{3D}$ &abs rel.  &$\text{time}$ \\
    \midrule
    (a) & SolvePoly \cite{bradski2000opencv} &53.5 &66.38 &0.078 &120  \\
    (b) & LTI & \textbf{53.4} & \textbf{66.38} &\textbf{0.078} &\textbf{23} \\
   \bottomrule
\end{tabular}
\label{tab:lut}  
\vspace{-0.1in}
\end{table}
\begin{table}[t]
\small
\centering
\caption{Comparison between initial structure design and lightweight structure design with FPNet on \textbf{FPD}. }
\begin{tabular}{c|c|cc}
    \toprule
	 & Method &$\text{time}$ & fps  \\
    \midrule
    (a) & w/o lightweight   & 180 & 5.5 \\
    (b) & w lightweight  & \textbf{23} & \textbf{43.5} \\
   \bottomrule
\end{tabular}
\label{tab:running time}  
\vspace{-0.2in}
\end{table}
\begin{table}[t]
\small
\centering
\caption{Comparison between two bbox center settings.
2D$\Rightarrow$3D indicates the offset from 2D center to the projected 3D center and vice versa.
}
\begin{tabular}{c|c|ccc}
    \toprule
	 & Method & $AP_{2D}$ & $AP_{3D}$ &abs rel.    \\
    \midrule
    (a) & 2D center + 2D$\Rightarrow$3D  & 48.2 & 60.25 &0.078 \\
    (b) & 3D proj. center + 3D$\Rightarrow$2D    & \textbf{53.4} & \textbf{66.38} &\textbf{0.078} \\
   \bottomrule
\end{tabular}
\label{tab:center}  
\vspace{-0.1in}
\end{table}

For 2D object detection on \textbf{FPD}, our FPNet outperforms both anchor-based pipeline RetinaNet \cite{article3} and the anchor-free pipeline CenterNet \cite{duan2019centernet}, KM3D \cite{li2021monocular} and MonoCon \cite{liu2021learning}.
For anchor-based RetinaNet \cite{article3}, our \textbf{FPD} contains abundant objects with various and large-angle poses, so presetting anchors maybe not fit with our dataset.
Our FPNet also follows the center-based structure \cite{duan2019centernet, li2021monocular, liu2021learning}, but these baselines \cite{duan2019centernet, li2021monocular, liu2021learning} directly predict 2D bbox center, which reveals greater challenge, because our 2D labels come from projected 3D labels instead of directly labeling on 2D images, so our 2D bbox centers may not indicate the semantics of 2D object centers.
Projected 3D center prediction with offset estimation from 3D center to 2D center assists FPNet to create a more correct 2D bbox center, resulting in better $AP_{2D}$ and $AR_{2D}$, also greater $AP_{3D}$ and $AP_{BEV}$, for precise 2D centers strengthen the other subtasks, like object center's depth and the heading angle.
Consequently, our FPNet also makes considerable 3D metrics improvement, compared with KM3D \cite{li2021monocular} and MonoCon \cite{liu2021learning}.
For dense depth estimation, the multi-task structure shares the perception knowledge, boosting the single depth estimation model, so our FPNet gains advancement.

\subsubsection{\textbf{Cross-dataset Evaluation}}
We validate our \textbf{FPD}'s generalization to other public datasets.
We train CenterNet \cite{duan2019centernet} object detection models and DRO \cite{gu2021dro} depth estimation model on two general datasets, COCO \cite{lin2014microsoft} and KITTI \cite{geiger2012kitti}, as shown in Table \ref{cross-dataset}. 
Firstly we train the models on our \textbf{FPD} (first row), then adopt the pre-trained models to finetune on another dataset (second row).
After finetuning, our \textbf{FPD} gains about 2\% to 4\% enhancement to the baselines directly trained on public datasets.
\textbf{FPD}'s various and diverse cases cover abundant autonomous driving scenes, which considerably lifts generalization ability.

\subsubsection{\textbf{Comparative Study for Lightweight Design}}
We conduct several comparative experiments for our targeted lightweight designs, where we manage to reduce the model's parameters, while still preserving the performance.

\textbf{Shared feature extractor}. Firstly, we explore the two changes with the shared feature extractor, as shown in Table \ref{lightweight1}.
\label{exp_share}
From (a), the initial setting of normal down-sampling radio 4 and complete layers performs best.
But from experiments (b) and (c), we find that only a tolerable and small performance drops when the input images are smaller (c) or remove some redundant layers (b).
So we finally adopt the settings (d) to achieve the most lightweight shared feature extractor.

\textbf{Other auxiliary monocular contexts}. Secondly, we investigate the effects of leaving out other auxiliary monocular contexts.
The other auxiliary monocular predictions consist of eight projected keypoints heatmap and offset vectors estimation, the quantization residual estimation with bbox center.
The experiment is exhibited in Table \ref{lightweight2}.
From settings (a) and (b), we see the additional keypoints tasks assist the 2D and 3D object detection, but consume more time. 
From settings (a) and (c), quantization residual estimation seems not much essential with little performance improvement.
We discard both settings in (d), and the operation makes the progress in real-time aspect, still with great results.

\begin{figure*}[!t]
    \centering
    \includegraphics[width=1\textwidth]{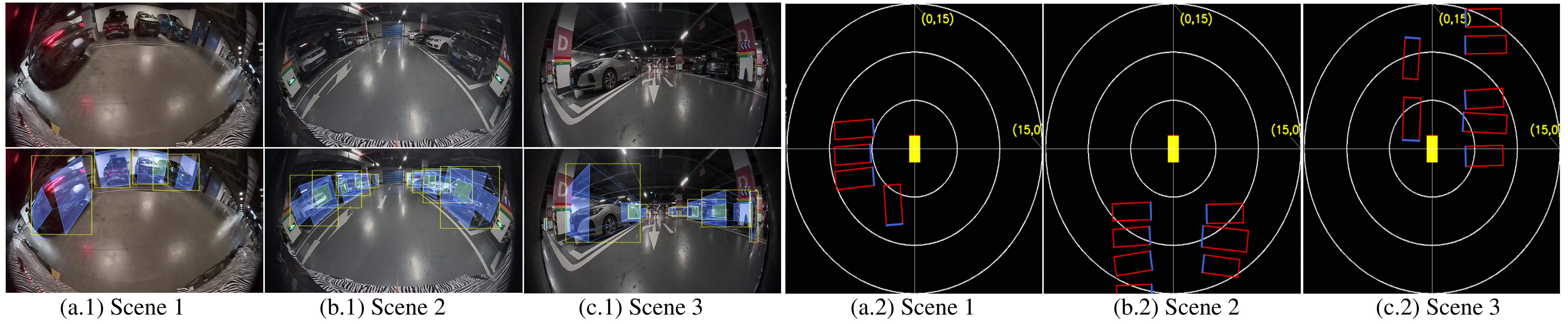}
    \caption{More qualitative results with more objects and directions on \textbf{FPD}. 
    (a) the input images; (b) the 2D object detection results (yellow bbox) and projected 3D object detection results (blue bbox); (c) BEV visualization from 3D results where the red bounding box denotes the predicted results and the blue segment denotes the objects' head.}
    \label{fig:quality2}
    \vspace{-0.2in}
\end{figure*}
\textbf{Look-up table indexing}. 
Thirdly, we discuss the look-up table indexing in the fisheye distortion module, as shown in Table \ref{tab:lut}.
Before our lightweight design, we adopt the solvePoly function from Opencv library \cite{bradski2000opencv}, which costs much unaffordable time for real-time application (setting (a)).   
Apparently, the indexing approach (b) saves more time than (a) without disturbing the detection scores.

\textbf{Time summary}. 
Finally, we perform time consumption on the hardware platform, and \yz{Table} \ref{tab:running time} reveals the inference time with lightweight design and without lightweight design.
From (a) and (b), lightweight design makes much sense, for we carefully develop advanced solutions to replace the elder's time-intensive operations.


\subsubsection{\textbf{Comparison with different bbox center settings}}
\label{exp_bbox}
Different from \cite{li2021monocular,liu2021learning}, we directly predict the projected 3D bbox center heatmap, instead of the 2D bbox center heatmap and offset from the projected 3D center to 2D center.
Table \ref{tab:center} exhibits that the direct 3D bbox center regression performs better.
We blame that we produce 2D bbox ground truth from outer bounding rectangle of projected 3D labels, but this 2D bbox label's center may not reflect the actual object's 2D image center, such as Motorbike and Traffic\_cones demos in Figure \ref{fig:objects_summary}, which alleviates the errors of 3D detection, especially the heading angle.
While \cite{li2021monocular,liu2021learning} utilize the correct 2D bbox center from correct 2D bbox labels.
Consequently, we adopt the proposal of directly regressing 3D bbox center and offset from 2D center to the projected 3D center.

\vspace{-0.1in}
\subsection{Qualitative Results}
\begin{figure}[!t]
    \centering
    \includegraphics[width=0.49\textwidth]{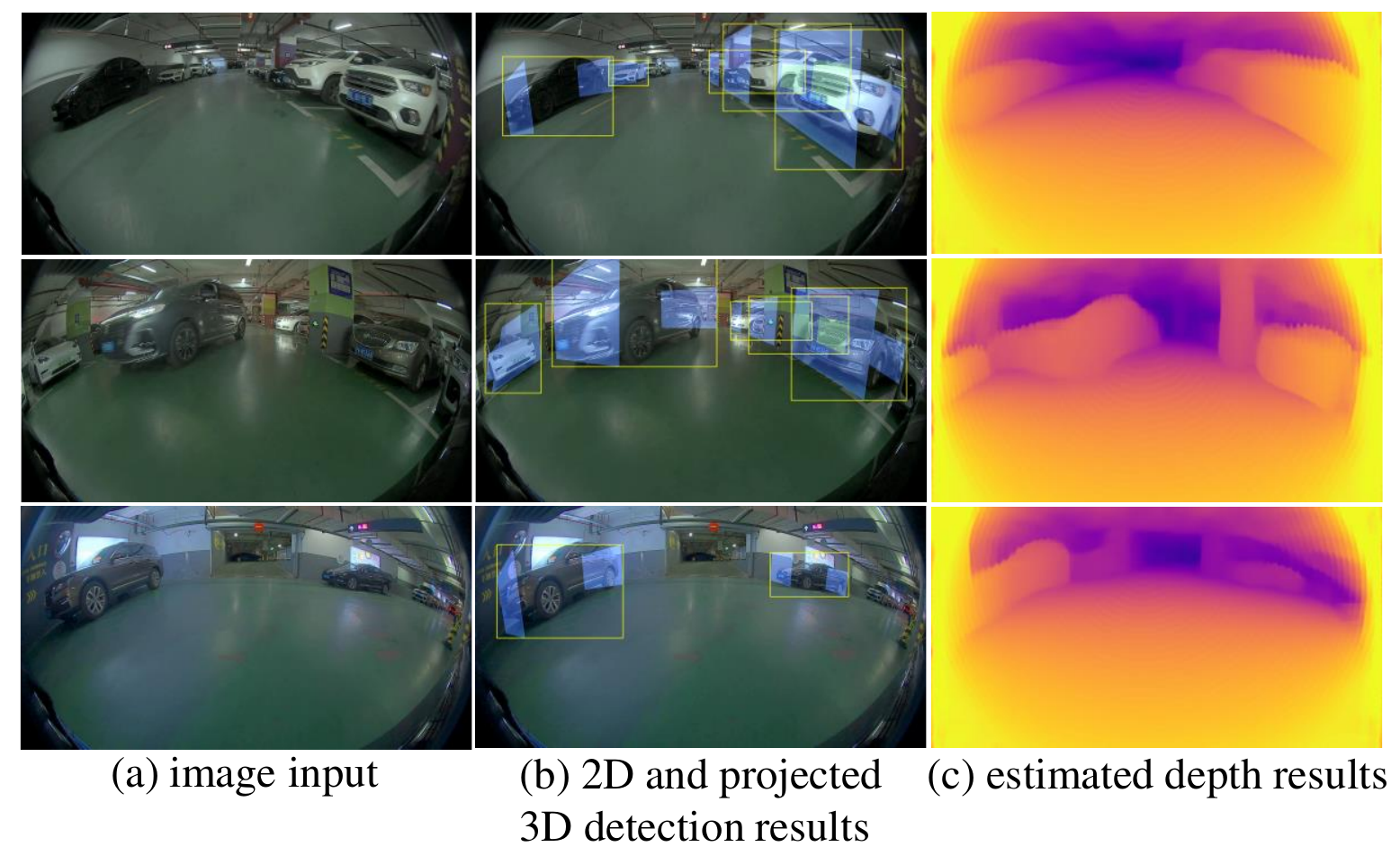}
    \vspace{-0.2in}
    \caption{Qualitative results with several valet parking scenes on \textbf{FPD}. 
    (a) the input images; (b) the 2D object detection results (yellow bbox) and projected 3D object detection results (blue bbox); (c) estimated depth results.
    }
    \label{fig:quality1}
    \vspace{-0.1in}
\end{figure}

We provide qualitative examples on our \textbf{FPD}, as shown in Figure \ref{fig:quality1} and Figure \ref{fig:quality2}.
Figure \ref{fig:quality1} displays some qualitative results with several valet parking scenes.
(b) denotes the 2D and 3D projected results and (c) denotes the estimation dense depth.
Our multi-task network satisfies the real-time perception requirement in different valet parking scenes.


Figure \ref{fig:quality2} demonstrates some visualization results with more objects and other directions.
Notably, (c) denotes the BEV visualization from 3D detection results.
From Figure \ref{fig:quality2}, our network achieves good perception performance for crowded objects in different directions.  


\vspace{-0.1in}
\section{Conclusion}
This article presents a new large-scale fisheye dataset, the \textbf{F}isheye \textbf{P}arking \textbf{D}ataset (\textbf{FPD}).
By providing a diversity of surround-view parking scenes, the proposed dataset aims to assist the industry in constructing a more secure advanced driving assistance system in parking lots. 
Moreover, we provide our real-time multi-task \textbf{F}isheye \textbf{P}erception \textbf{N}etwork (\textbf{FPNet}), for enhancing fisheye distortion performance and various lightweight designs. 
Extensive experiments on \textbf{FPD} validate the effectiveness of our \textbf{FPNet}. 
However, \textbf{FPD} has a lot of capacity for development, including how to strengthen more data diversity, simplify our approach, and deal with sustainably increasing data and the potential for diverse vision tasks. 
Nevertheless, we expect \textbf{FPD} to inspire more relevant research and promote the performance of surround-view perception under parking scenes.

\yz{
In the future, we will further explore the surround-view fisheye BEV perception for valet
parking, in the following aspects. (1) We will explore the LiDAR-camera fused perception since the LiDAR point cloud provides more 3D information.
(2) We will explore the lightweight image features extractor to achieve more robust visual features.
(3) We will design specific techniques to augment the distorted fisheye images to enhance the generalization ability.
(4) We will apply our surround-view fisheye BEV perception and dataset  refer to other tasks, such as obstacle detection.
}

\bibliographystyle{IEEEtran}
\bibliography{ref}

\end{document}